# A Patient-Doctor-NLP-System to contest inequality for less privileged

SUBRIT DIKSHIT[1*], RITU TIWARI[1], PRIYANK JAIN[1]

[1] Department of Computer Science and Engineering, Indian Institute of Information Technology, Pune, India

[*] Corresponding Author

E-mail: subrit@gmail.com[1*], ritu@iiitp.ac.in[1], priyank@iiitp.ac.in[1]

**Abstract.** Transfer Learning (TL) has accelerated the rapid development and availability of large language models (LLMs) for mainstream natural language processing (NLP) use cases. However, training and deploying such gigantic LLMs in resource-constrained, real-world healthcare situations remains challenging. This study addresses the limited support available to visually impaired users and speakers of low-resource languages such as Hindi who require medical assistance in rural environments. We propose PDFTEMRA (Performant Distilled Frequency Transformer Ensemble Model with Random Activations), a compact transformer-based architecture that integrates model distillation, frequency-domain modulation, ensemble learning, and randomized activation patterns to reduce computational cost while preserving language understanding performance. The model is trained and evaluated on medical question-answering and consultation datasets tailored to Hindi and accessibility scenarios, and its performance is compared against standard NLP state-of-the-art model baselines. Results demonstrate that PDFTEMRA achieves comparable performance with substantially lower computational requirements, indicating its suitability for accessible, inclusive, low-resource medical NLP applications.

## 1 Introduction

This is the era of LLMs, such as BERT [1], BART [2], T5 [3], GPT2 [4], Bloom [5], Llama [6] and many more which are readily and easily accessible to everyone. Many of these LLMs underlying use attention-based-transformers [8]. Transformers, have achieved great accomplishment in realizing AI and NLP based tasks. GPT2, a transformer decoder-based-network turned out as few of highest cited research. Academicians and researchers have strained to overcome restrictions of transformers by introducing finer improved networks. Majority work exploited similar time-domain improvements and minimum use of forward-thinking scientific notions, such as,

Modulation [9], Frequency Domain Analysis [10], Fourier Transform [11] as in FNET [12] or Hartely Transform [13] to further heighten the network performance. Some researchers tried concepts like distillation as anchored in DistilBERT [14] to improve network efficiency utilizing Teacher-Student network to get similar performances of large models with smaller and smarter architectures. WISHPER [15], a multilingual speech-to-text network predicts text transcripts based on 0.68 million hours of training and generalizes nicely on benchmarks. MMS [16], a transformer-based-speech-model scales to 1000+ languages and is built on wav2vec 2.0 base. M2M-VITS [17], with M2M100 baseline, is a multilingual transformer-based-network utilized in multilingual transformations that can translate in 100 languages. However, these models are humongous and cannot be trained from scratch owing to accompanying costs. Training these models takes from many days to months on multiple GPUs together. However, to reduce this training time from scratch, techniques like, fine-tuning, quantization, compression and adaptation are utilized. LORA [18], known as Low-Rank Adaptation is lightweight training procedure to lessen count of trainable parameters by introducing a minor count of novel weights into the network and are trained to produce smaller model. Use of AdaNorm [19], in PDFTEMRA, an outperforming layer normalizing technique results faster and memory-efficient model architecture.

Activations are useful in neural networks resulting smarter inferences. TanH, as utilized in LSTM [20] is chosen above sigmoid for superior non-linearity but suffers with vanishing gradients that is better dealt by ReLU [21]. ReLUs, known as Rectified Linear Units are linear on non-negative measurements and 0 for negatives resulting improved non-linear knowledge retention. ELU [22], known as Exponential Linear Unit (ELU) induces exponential activation learning that has negatives near to 0's consequential inferior computational complications. PreLU [23] activations have negative values as well but they ensure lesser clatter deactivations. LeakyReLU [24], known as Leaky Rectified Linear Unit is framed on ReLU adding shorter gradients for non-positive values rather than flat gradients. SELU [25], known as Scaled Exponential Linear Unit encourage self-normalization. CELU [26] is good for raising deeper networks by quality of non-vanishing-gradients. MISH [27], is self-regularized non-monotonic outperforms LeakyReLU on YOLOv4 [28] and ReLU on ResNet-50 [29]. SILU [30], known as Sigmoid Linear Unit is calculated with sigmoid increased with inputs. GELU [31], known as Gaussian Error Linear Unit is standard and smoother than ReLU and utilized in BERT, GPT3 [32] and many other known transformers.

Evaluation Metrics are used to validate our network performance. WER [33], known as Word error rate metric validates speech recognition/translation performance and ranges between 0 to 1, where 0 indicates exactly identical and 1 absolutely dissimilar outcome. BLEU [34], known as Bilingual evaluation evaluates machine-translated transcripts and claims higher relationship with human based decisions. BLEU score ranges between 0 to 1, where values nearing 1 represent resemblances and 0 non-resemblances.

## 1.1 Highlights

**Purpose:** The drive of this study is to develop and validate PDFTEMRA (Performant Distilled Frequency Transformer Ensemble Model with Random Activations), a compact transformer-based network engineered to offer accessible medical natural language processing capabilities for visually impaired users and speakers of low-resource languages like Hindi in resource-constrained rural healthcare settings, while preserving performance analogous to state-of-the-art models but with significantly reduced computational necessities.

**Contributions:** PDFTEMRA solution contributions are discussed below:

- **PDFTEMRA Implementation** – PDFTEMRA (Performant Distilled Frequency Transformer Ensemble Model with Random Activations) demonstrates successful use of Transfer Learning (TL) on the custom medical dataset, PADT (PDFTEMRA Dataset).
- **Instrumenting with Model Distillation (MD)** – Model Distillation (MD) technique is grounded on Knowledge Distillation (KD) that replaces, Teacher with Distributor (D) and Student with Consumer (C) model. There could be multiple (Ds) based on the needs and required tasks and (C) will mimic (D) based on task-in-hand. Implementation selects GPT2 as (D), to teach, PDFTEMRA as (C) model which achieve excellent results.
- **Validate use of novel Layer Transformation Technique** – This work, demonstrates successful usage of Frequency Modulation techniques instead of Attention, via, Hartley transform to realize cheaper and faster inference.
- **Establish use of novel Efficient Normalization Technique** – The network utilizes AdaNorm instead of the casual layer normalization that enhances our network performance even further.
- **Demonstrate use of novel Activation Ensemble (AE)** – Practice of Activation Ensemble (AE), built with the multiple smart activations act more efficiently and establish as very performant layer using ELU, LeakyReLU, PReLU, ReLU, SELU, CELU, Mish, Tanh, SiLU, and GELU.
- **Demonstrate use of novel Network Ensemble (NE) for Implementation of Patient-Doctor-NLP-System** – Network Ensemble (NE), utilizes multiple models by sandwiching together. This work creates WISPHER_PDFTEMRA_MMS_VITS ensemble that is the combination of PRE-PROCESSOR_PROCESSOR_POST-PROCESSOR ensemble architecture, where, WISPHER is the PRE-PROCESSOR, PDFTEMRA is the PROCESSOR, MMS and VITS are the POST-PROCESSORs.
- **PADT (PDFTEMRA Dataset)** – A custom high quality medical dataset, PADT (PDFTEMRA Dataset) is created for training and fine-tuning our PDFTEMRA network to demonstrate successful implementation of Patient -Doctor-NLP-System.

**Challenges:** Next challenges towards implementation of Patient-Doctor-NLP-System for Hindi speaking least privileged or visually impaired audience are next stated below:

- **Motivational Encounter** – Race towards capitalism, many are eager to contribute towards expanses with huge financial gains. However, this work ideates to solve an issue that paralyzes a segment of our society and has huge impact. The implementation goal is to help the less privileged and reduce social inequality by assisting blind or unschooled Hindi speaking patients seeking medicine support. During this research, it was understood that, less privileged are more often concentrated towards rural areas and doctors in urban, probably, with reason to have good lifestyle and ease of living in bigger cities. This gap is expanding day-by-day and hence the work tries to shrink this discontinuity.

- **Software and Hardware Limitation** – The research had access to open-source software resources and limited hardware as we lacked funding support. The open-source models used were run on RTX 3070 8GB GPU, Intel 4.7 GHz 20 Core CPU and ram with 32 GB size.
- **Dataset Availability & Quality** – The PADT dataset is baselined on freely available datasets and web-scraped-data. Thus, the work suits academic inferences and not fully compatible with commercial product implementation. As there is a huge commercial market, wherein, the work can be implemented and while suggest to the readers is to strengthen the datasets with robust doctor-based-expert-opinions for commercial product or solution implementations.
- **Model Bias** – The work utilizes open-source networks, for reference, Distributor (D), Pre-Processor and Post-Processor. Reader should know that these models may have induced biases which might get passed-on to Consumer (C) model. Therefore, for any commercial product or commercial solution implementation based on the research work, model biases should be handled effectively.

**Findings:** The work demonstrates that PDFTEMRA achieves performance comparable to standard state-of-the-art NLP networks on medical question-answering and consultation PADT dataset tailored to Hindi and accessibility scenarios, while requiring substantially lower computational resources. This indicates that the integration of model distillation, frequency-domain modulation, ensemble learning, and randomized activation patterns successfully reduces computational cost without compromising language understanding performance, making the model suitable for deployment in resource-constrained rural healthcare environments where visually impaired users and speakers of low-resource languages require medical assistance.

The entire paper is divided into VI sections. Section II reflects the associated research. Section III talks about the methodology and approach behind the network. Section IV will be discussing the results of the experiment. We deliver our inferences in Section V and references are embedded in Section VI.

## 2 Related Works

The reviewed literatures and related works are discussed in this section. However, this research work is unique as the most of other the researchers concentrated on the use of computer vision rather than NLP techniques while trying to contest analogous problems while offering the enablement for the less privileged.

Khan et al. (2020) [35] derived that blindness might cause individual from knowledge acquisition about environment and offer a Raspberry Pi-3 prototype for object recognition and obstacle avoidance, using AI model, enabling effortlessness navigation for impaired visually.

Smith et al. (2021) [36] presented work that expresses on experiences where AI do assist and frustrates in day-to-day tasks. Yet, in totality, author states that AI based experiences are optimistic.

Sharma et al. (2020) [37] presented sixth sense cellphone app and website. Cellphone app for object detection and website for text-to-speech recognition or translation. However, shortfall for the experiment was dependency on camera picture quality.

Khan et al. (2022) [38] presented a review-based results that contribute on AI chat-based health agents, knowledge gaps and issues.

Leerang et al. (2024) [39] also presented a review that concentrated on user research, and experience and component evaluation. Results depict four strategies. One, personalized user interface and user experience (UI/ UX). Two, important functionality to back communications. Three, behavioral induction system and four, vigorous functional scheme guaranteeing care and conviction.

Qureshi et al. (2021) [40] presented their paper which focused on real time AI image recognition and obstacle detection to support visually disabled for self-sufficiently location steering while the proposed system was not completed.

Muqbali et al. (2020) [41] presented the project based on Open CV and Python whose objective was to support blind in daily activities via smart device. Vision AI based output device notice faces, colors and obstacles outputting vibration or audio alert to support visually impaired.

## 3 Methodology

Herein below, the stepwise overview of proposed work methodology is offered along with the vital philosophies and utilized guiding forces which are next defined subsequently, thereafter.

- Define the problem, solution and architecture for the network.
- Prepare and process the Dataset.
- Train the network Processor.
- Finetune the Pre-Processor and the Post-Processor.
- Optimize, verify and validate overall network and the performance.
- Execute and generate the inferences for this work.

**Overall Implementation:** The bird-eye view of the comprehensive implementation is revealed below in the Figure 1.

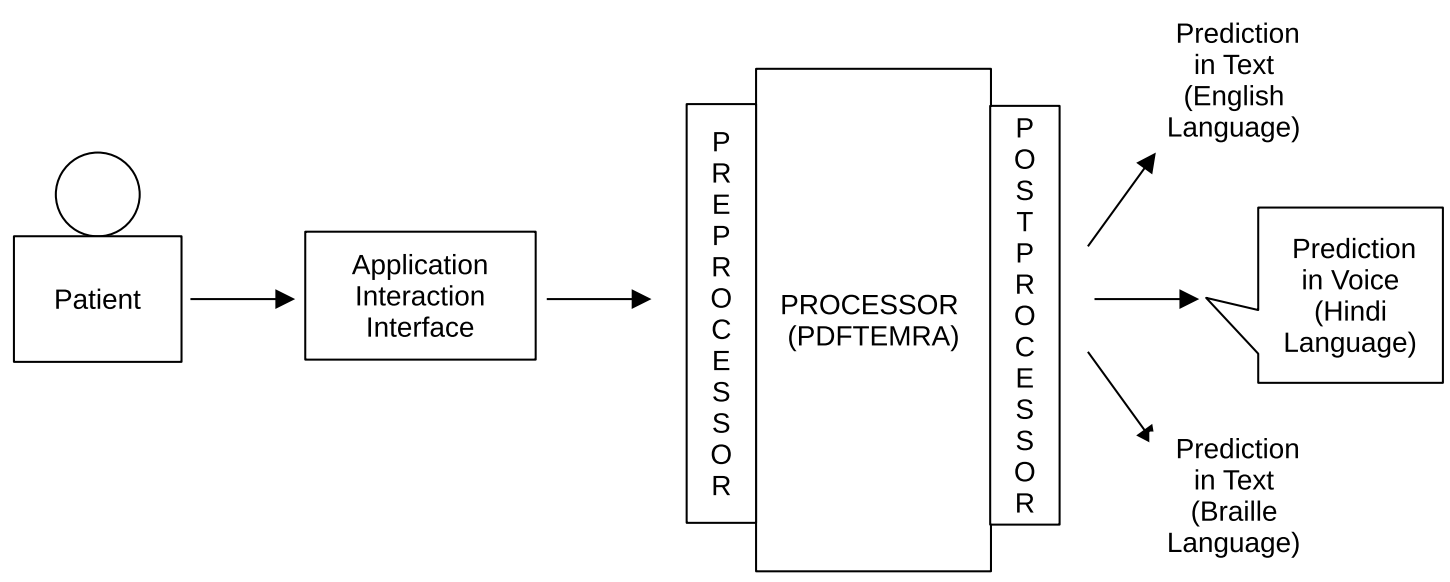

**Fig. 1.** Overall Implementation with Inference.

Our patient, less privileged one (i.e. either cannot read/write or blind and speaks Hindi/English) interacts with Application via voice prompts (speaks native language Hindi/English). The local webserver-based Website/Android/iOS Apps are used as the interaction layer. The ensembled PDFTEMRA network consists of the [PRE-PROCESSOR + PROCESSOR + POST-PROCESSOR]. The PRE-PROCESSOR, processes voice prompt inputs in the Native (i.e. Hindi) or English language and feeds to the PROCESSOR. The PROCESSOR, digests this input and generates the medication predictions. The POST-PROCESSOR, feeds to PROCESSOR's yield and generates plays the medication speech-based-output back in Native (i.e. Hindi), displays medication on screen in text (i.e. English) and Braille language. This makes network output more usable that could be printed or e-mailed to an expert doctor for second opinion. Next, let us discuss utilized vital philosophies and guiding forces.

**Attention, Fourier and Hartley Transform:** Modulation**,** an impression that eases information quality missingness issue and is achieved via adding extra participation in fully connected layer (FLL). Transfer function of FLL and Modulated FLL (MFLL) of $L_{in}$ input to layer is given by:

$$L_o = f(w \,.\, L_{in} + b), w^m = k^W(\theta), \;\; b^m = k^b(\theta) \tag{1}$$

Where, $\theta$ is the modulating signal, f is non-linearity function, g is the function of perceptron with multiple layers, $w$ and $b$ weight and bias. Thus $L_o$ is given by:

$$L_o = f(L_{in} \,.\, k^W(\theta)) \tag{2}$$

Fourier transform (FT) and Hartley transform (HT) are employed to the layers that replace attention layer in our network. The Distributor (D) model, follow standard attention (A) scores and is equated by:

$$\text{A}\ (Q, K, V) = \text{S}\ (\frac{QK^T}{\sqrt{d_k}})\text{V} \tag{3}$$

Where query *Q*, key *K,* value *V*, and S softmax are the representations. For input stream $I^j$, where 0≤ j, $k \leq N-1$ and N is length, discrete Fourier representation summation of $I^n$ is given by:

$$I^t = \sum_{j=0}^{N-1} I_n \left(exp^{\frac{-(2\pi itn)}{N}}\right) \quad (4)$$

Hartley transform gives similar performance to FT and that transmutes real input to yield is given by:

$$HT = R[FT] - J[FT] \quad (5)$$

**Multiple Activation Ensemble:** In an ensemble, each neuron activation is given by:

$$O_k = M(a^p) \quad (6)$$

Where, $a$ being the original activation function while additional activations allows the model to learn and adapt better.

**AdaNorm:** AdaNorm ($A_n$) for *v* input vector it is given by:

$$A_n = K\,(1 - lt) \circ t,\;\; t = A - \mu/\sigma,\;\; \mu = 1/C\,(\sum_{i=1}^{C} a^i),$$

$$\sigma = \sqrt{1/C\,(\sum_{i=1}^{C}(a^i - \mu)^2)} \quad (7)$$

Where, *K* is hyper-parameter, ° dot product, k is 0.01, $t = (t_1, \ldots, t_C)$, A the input sequence, C is dimension of t, $a^i$ is ith element, $\mu$ and $\sigma$ are mean and variance.

**Adaptation:** Adaptation substitutes fine tuning via additional adapter layers to lower projection of the original higher dimensional features, while adds non-linearity and then upscales back to original projection. This is achieved by introducing $\delta_i$ work type to pre trained network parameters $\delta$, where ($1 \leq i \leq M$) for M tasks. Adaptor, limits number of available inputs to transfer representations effectively from older to new ones and thus easing damage of information. This fused objective is given by:

$$Min^{J^i}(F_D)^{-1} \sum_{k=1}^{D}(-p(J'a_i, \beta') + p(J, a_i, \beta))^2 \quad (8)$$

Where, $J^i$is combined network parameters, $F_D$is unlabeled drill information, loss of distillation is logit computer change by current model $p(J, a_i, \beta)$ versus consolidated model $p(J'a_i, \beta')$ and once $F_D$ is skilled $\beta$is substituted by $\beta'$.

**Transfer Learning:** Transfer learning is practiced using OPENAI/WHISPER-LARGE-V3 as the PRE-PROCESSOR, FACEBOOK/M2M100_418M and FACEBOOK/MMS-TTS as the POST-PROCESSORs. These efficient models are selected as being open-source, and allowing easy fine-tunning that suits this study. On the other hand, creating a similar fresh model is not beneficial, as it is not financially feasible with limited software and hardware at disposal. Re-doing WHISPER-LARGE-V3 with 1.55 billion parameters will cost around $218700 million/year; while M2M100_418M costing $491000 million/year, MMS-TTS along with 1 billion parameters would be costing $153,090 million/year approximately.

**Model Distillation:** The practice of Model Distillation (MD) via Distributor (D) and Consumer (C) networks for the defined work is shown in the Figure 2 below.

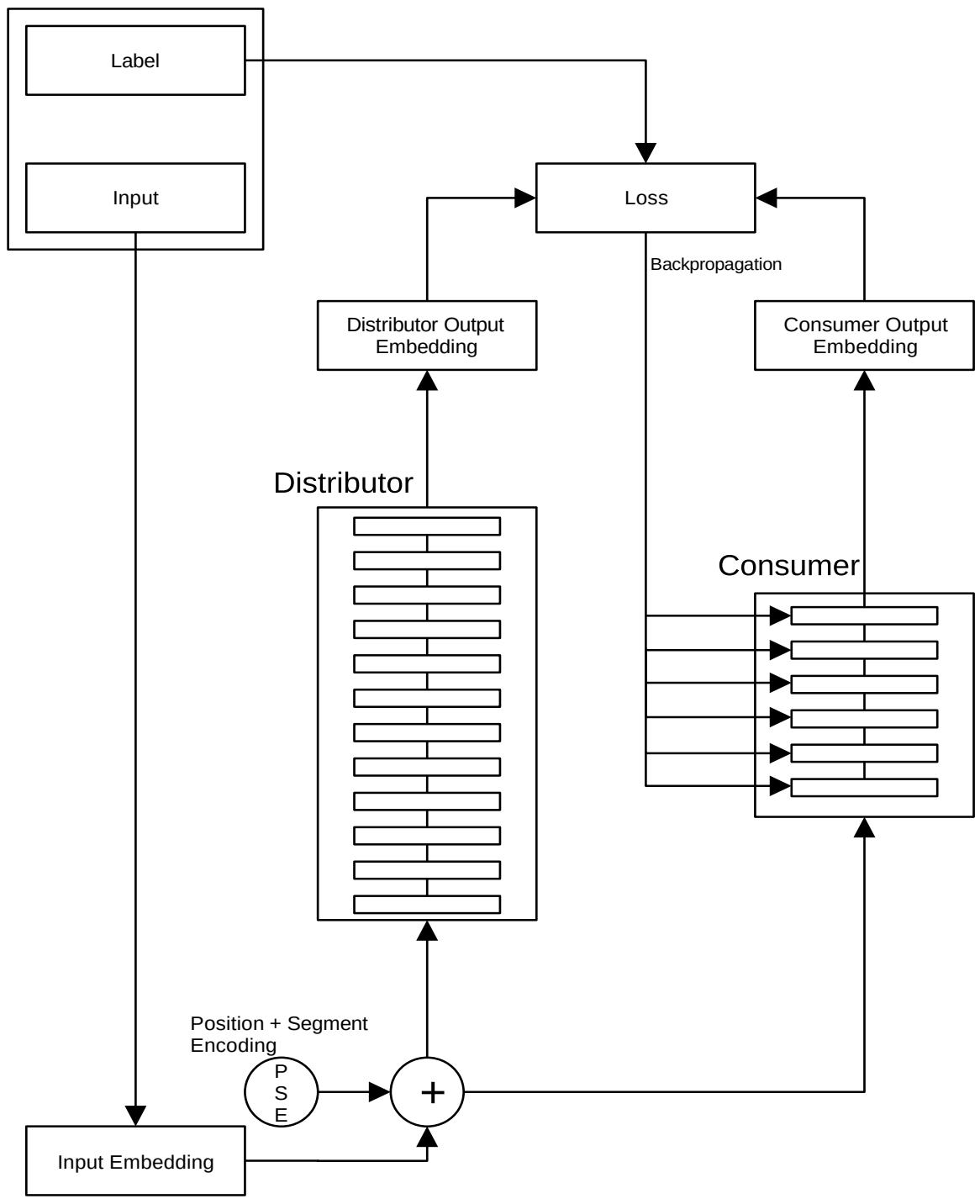

**Fig. 2.** Model Distillation (MD) along with Distributor (D) & Consumer (C).

OPENAI-GPT2 acts as the choice of Distributor (D) and the PDFTEMRA as (C). PDFTEMRA ability to grasp the knowledge is founded on how virtuous is (D) and (C)'s self-learning capabilities. Given $D_i$ and $G_i$ are probabilities for $i^{th}$ element, the training distillation loss is given by:

$$L_D = \sum_i G_i * \log D_i \tag{9}$$

PDFTEMRA takes input and adds segment and position embedding to it. This is passed on to HT layer with different activation ensemble for each head. Output is summed passed to AdaNorm layers for input normalization along with dropout. This is followed up by FeedForward and AdaNorm layer again. Finally, output is retrieved via Linear layer followed by SoftMax. Detailed architecture is shown in Figure 3 along with visual contrast from original transformer setup.

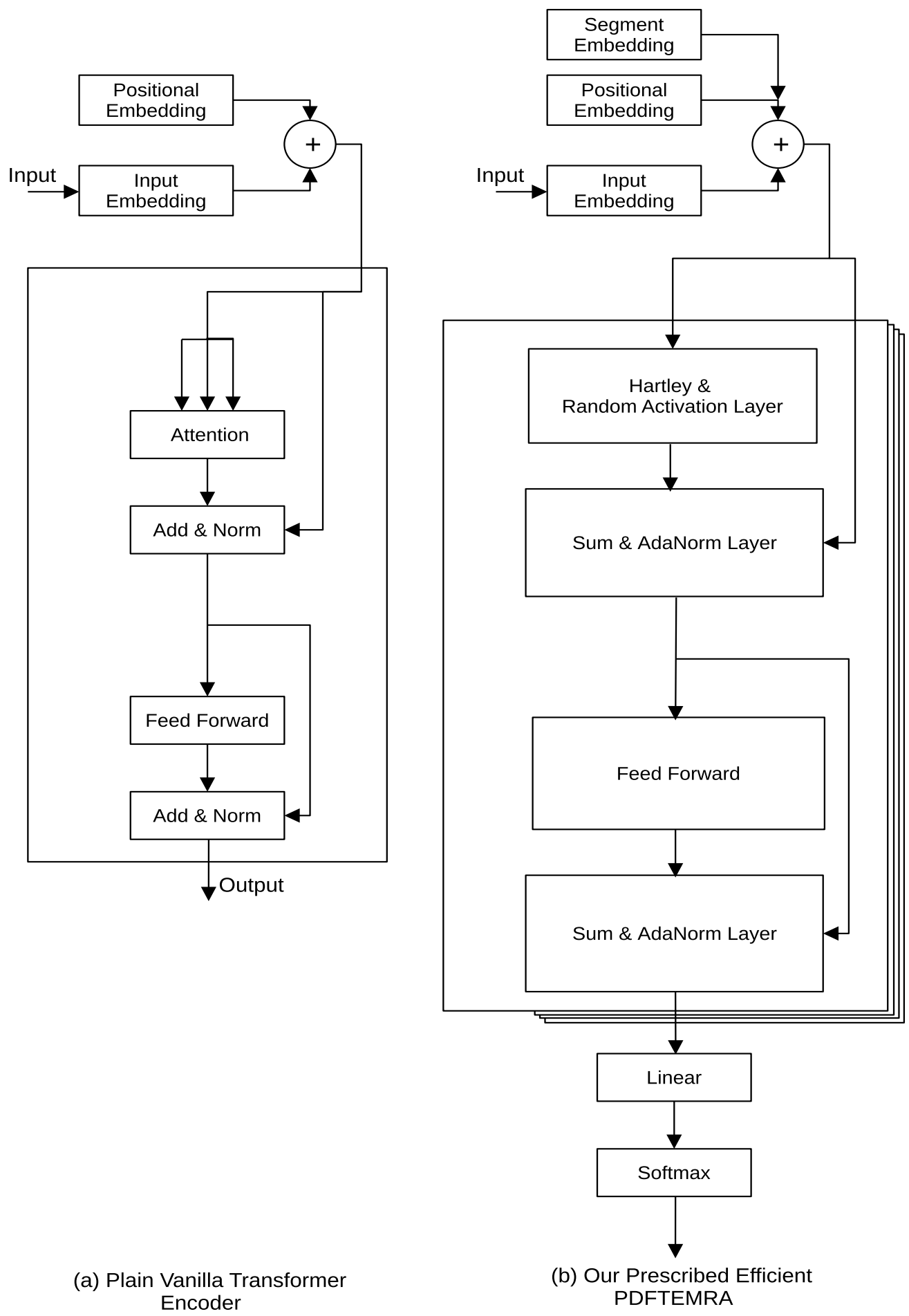

**Fig. 3.** Original Transformer Vs PDFTEMRA architecture.

## 4 Experiment & Results

This section discusses the setup, experiment and obtained inferences.

**Datasets:** CFILT/IITB-ENGLISH-HINDI dataset for English-Hindi translation helps fine-tuning of the network. The IIT Bombay English-Hindi corpus consists of English-Hindi text pairs. MOZILLA-FOUNDATION/COMMON_VOICE_11_0 dataset composed of acoustic mp3 and matching text pairs (Train: 24210 + Validation: 16413) hours is used for speech training and tunning.

Training of the PDFTEMRA model on PADT dataset entails to English-Hindi language pairs (Train: 122000 + Validation: 2490 + Test: 5010) samples. PADT dataset is grounded using the KAGGLE/DATASETS/MOHNEESH7/INDIAN-MEDICINE-DATA of size 354.1 MB consisting of the Indian medicines along with KAGGLE/DATASETS/SINGHNAVJOT2062001/11000-MEDICINE-DETAILS dataset that consists 11000 Indian medicines manually web scraping records from 1MG website.

PADT dataset can be requested from the author and available at:

**Dataset - PADT:**

https://drive.google.com/drive/folders/1aUeGJ29DvQ98RapGGoSrQCabaYFbMxuP?usp=drive_link

**Training:** Model Distillation MD of the PDFTEMRA network is achieved using PADT dataset for 25 epochs, learning rate at 0.001, temperature as 2, maximum sequence length at 512, number of head as 10, latent dim of 100 and embedded dim as 256.

The source code for PDFTEMRA can be requested from the author and is available at:

**Entire Source Code - PDFTEMRA:**

https://drive.google.com/drive/folders/1sKOB0hxs-PUvReQegH4DmvZy-5OVO5IX?usp=drive_link

Table 1 and Figure 4, displays Distributor (D) and Consumer (C) networks training and performance on the PADT dataset. Based on these results, during Model Distillation (MD) process, The network (C) takes more epochs to gain knowledge in comparision to plain vanilla (D) and (C) trainings. It is worth noting, that, the networks start to stablize around epoch 10, but, while testing vibrations are we obserbed in the distillation loss and accuracy making training the models until epoch 20 to 22 for better inferences.

**Table 1:** Model Distillation MD training and performance results for networks

| Epochs | D-TL | D-VL | DP-TL | DP-TDL | DP-VL | DP-VDL | C-TL | C-VL | D-TA | D-VA | DP-TA | DP-VA | C-TA | C-VA |
|---|---|---|---|---|---|---|---|---|---|---|---|---|---|---|
| 1 | 0.3666 | 0.3473 | 0.2113 | 0.0185 | 0.2192 | 0.0067 | 0.3704 | 0.3011 | 0.8336 | 0.8591 | 0.9315 | 0.9256 | 0.8303 | 0.8829 |
| 2 | 0.1566 | 0.1208 | 0.1045 | 0.0096 | 0.0919 | 0.009 | 0.1829 | 0.1149 | 0.9405 | 0.9562 | 0.9894 | 0.9894 | 0.9321 | 0.9585 |
| 3 | 0.066 | 0.0231 | 0.0793 | 0.0096 | 0.0712 | 0.0093 | 0.1072 | 0.0342 | 0.9764 | 0.9936 | 0.9992 | 0.9977 | 0.9616 | 0.9895 |
| 4 | 0.029 | 0.0021 | 0.0702 | 0.0095 | 0.0665 | 0.0093 | 0.068 | 0.0076 | 0.9902 | 0.9998 | 1 | 0.9993 | 0.9761 | 0.9986 |
| 5 | 1.67E-02 | 2.74E-04 | 6.76E-02 | 0.0095 | 6.54E-02 | 0.0093 | 0.0498 | 0.0021 | 0.9942 | 1 | 1.00E+00 | 1.00E+00 | 0.9832 | 0.9996 |
| 6 | 1.47E-02 | 8.65E-05 | 6.76E-02 | 0.0094 | 6.59E-02 | 0.0093 | 0.034 | 4.46E-04 | 0.9946 | 1 | 1.00E+00 | 9.99E-01 | 0.989 | 1 |
| 7 | 2.20E-02 | 4.18E-05 | 6.87E-02 | 0.0095 | 6.66E-02 | 0.0093 | 0.0284 | 1.09E-04 | 0.9926 | 1 | 1.00E+00 | 9.99E-01 | 0.9906 | 1 |
| 8 | 9.50E-03 | 2.29E-05 | 6.93E-02 | 0.0095 | 6.96E-02 | 0.0093 | 0.0278 | 5.15E-05 | 0.9967 | 1 | 1.00E+00 | 9.98E-01 | 0.9909 | 1 |
| 9 | 1.23E-02 | 1.42E-05 | 6.92E-02 | 0.0095 | 6.82E-02 | 0.0093 | 0.0193 | 2.75E-05 | 0.996 | 1 | 1.00E+00 | 9.99E-01 | 0.9938 | 1 |
| 10 | 1.32E-02 | 9.13E-06 | 6.81E-02 | 0.0094 | 6.62E-02 | 0.0093 | 0.0145 | 1.42E-05 | 0.9952 | 1 | 1.00E+00 | 9.99E-01 | 0.9957 | 1 |
| 11 | 8.60E-03 | 6.02E-06 | 6.77E-02 | 0.0094 | 6.74E-02 | 0.0093 | 0.0108 | 8.60E-06 | 0.9972 | 1 | 1.00E+00 | 9.99E-01 | 0.9967 | 1 |
| 12 | 4.20E-03 | 3.83E-06 | 6.75E-02 | 0.0094 | 6.54E-02 | 0.0093 | 0.0093 | 5.38E-06 | 0.9987 | 1 | 1.00E+00 | 9.99E-01 | 0.9968 | 1 |
| 13 | 8.80E-03 | 2.35E-06 | 6.66E-02 | 0.0094 | 6.57E-02 | 0.0093 | 0.01 | 3.41E-06 | 0.9969 | 1 | 1.00E+00 | 9.99E-01 | 0.997 | 1 |
| 14 | 1.45E-02 | 1.41E-06 | 6.71E-02 | 0.0094 | 6.56E-02 | 0.0093 | 0.0123 | 2.19E-06 | 0.9947 | 1 | 9.99E-01 | 9.99E-01 | 0.9961 | 1 |
| 15 | 7.10E-03 | 9.45E-07 | 6.57E-02 | 0.0094 | 6.42E-02 | 0.0093 | 0.004 | 1.38E-06 | 0.9976 | 1 | 1.00E+00 | 1.00E+00 | 0.9987 | 1 |
| 16 | 2.50E-03 | 6.35E-07 | 6.63E-02 | 0.0093 | 6.44E-02 | 0.0093 | 0.0123 | 0.002 | 0.999 | 1 | 1.00E+00 | 1.00E+00 | 0.9959 | 0.9994 |
| 17 | 4.00E-03 | 4.09E-07 | 6.57E-02 | 0.0093 | 6.59E-02 | 0.0093 | 0.0047 | 0.0238 | 0.9989 | 1 | 1.00E+00 | 9.99E-01 | 0.9988 | 0.9923 |
| 18 | 7.70E-03 | 2.83E-07 | 6.48E-02 | 0.0093 | 6.53E-02 | 0.0093 | 0.0039 | 0.0013 | 0.9972 | 1 | 1.00E+00 | 9.99E-01 | 0.999 | 0.9995 |
| 19 | 5.20E-03 | 1.77E-07 | 6.60E-02 | 0.0093 | 6.49E-02 | 0.0093 | 0.0061 | 9.58E-05 | 0.9982 | 1 | 9.99E-01 | 1.00E+00 | 0.9983 | 1 |
| 20 | 3.80E-03 | 1.11E-07 | 6.61E-02 | 0.0093 | 6.57E-02 | 0.0093 | 0.0058 | 3.09E-05 | 0.9985 | 1 | 9.99E-01 | 9.99E-01 | 0.9983 | 1 |
| 21 | 4.20E-03 | 7.58E-08 | 6.56E-02 | 0.0093 | 6.61E-02 | 0.0093 | 0.0032 | 1.90E-05 | 0.9983 | 1 | 9.99E-01 | 9.99E-01 | 0.999 | 1 |
| 22 | 4.00E-03 | 4.77E-08 | 6.50E-02 | 0.0093 | 6.58E-02 | 0.0093 | 7.02E-04 | 1.25E-05 | 0.9986 | 1 | 1.00E+00 | 9.99E-01 | 0.9998 | 1 |
| 23 | 1.60E-03 | 3.42E-08 | 6.57E-02 | 0.0093 | 6.40E-02 | 0.0093 | 0.0051 | 8.27E-06 | 0.9994 | 1 | 9.99E-01 | 1.00E+00 | 0.9982 | 1 |
| 24 | 5.80E-03 | 2.32E-08 | 6.46E-02 | 0.0093 | 6.45E-02 | 0.0093 | 0.0057 | 5.49E-06 | 0.9982 | 1 | 1.00E+00 | 1.00E+00 | 0.9982 | 1 |
| 25 | 8.30E-03 | 1.72E-08 | 6.41E-02 | 0.0093 | 6.42E-02 | 0.0093 | 0.0047 | 3.66E-06 | 0.9969 | 1 | 1.00E+00 | 1.00E+00 | 0.9984 | 1 |

## Legends - Table 1:

D-TL as Distributor-Training Loss, D-VL as Distributor-Validation Loss, DP-TL as Distillation Process-Training Loss, DP-TDL as Distillation Process-Training Distillation Loss, DP-VL as Distillation Process-Validation Loss, DP-VDL as Distillation Process-Validation Distillation Loss, C-TL as Consumer-Training Loss, C-VL as Consumer-Validation Loss, D-TA as Distributor-Training Accuracy, D-VA as Distributor-Validation Accuracy, DP-TA as Distillation Process-Training Accuracy, DP-VA as Distillation Process-Validation Accuracy, C-TA as Consumer-Training Accuracy, C-VA as Consumer-Validation Accuracy.

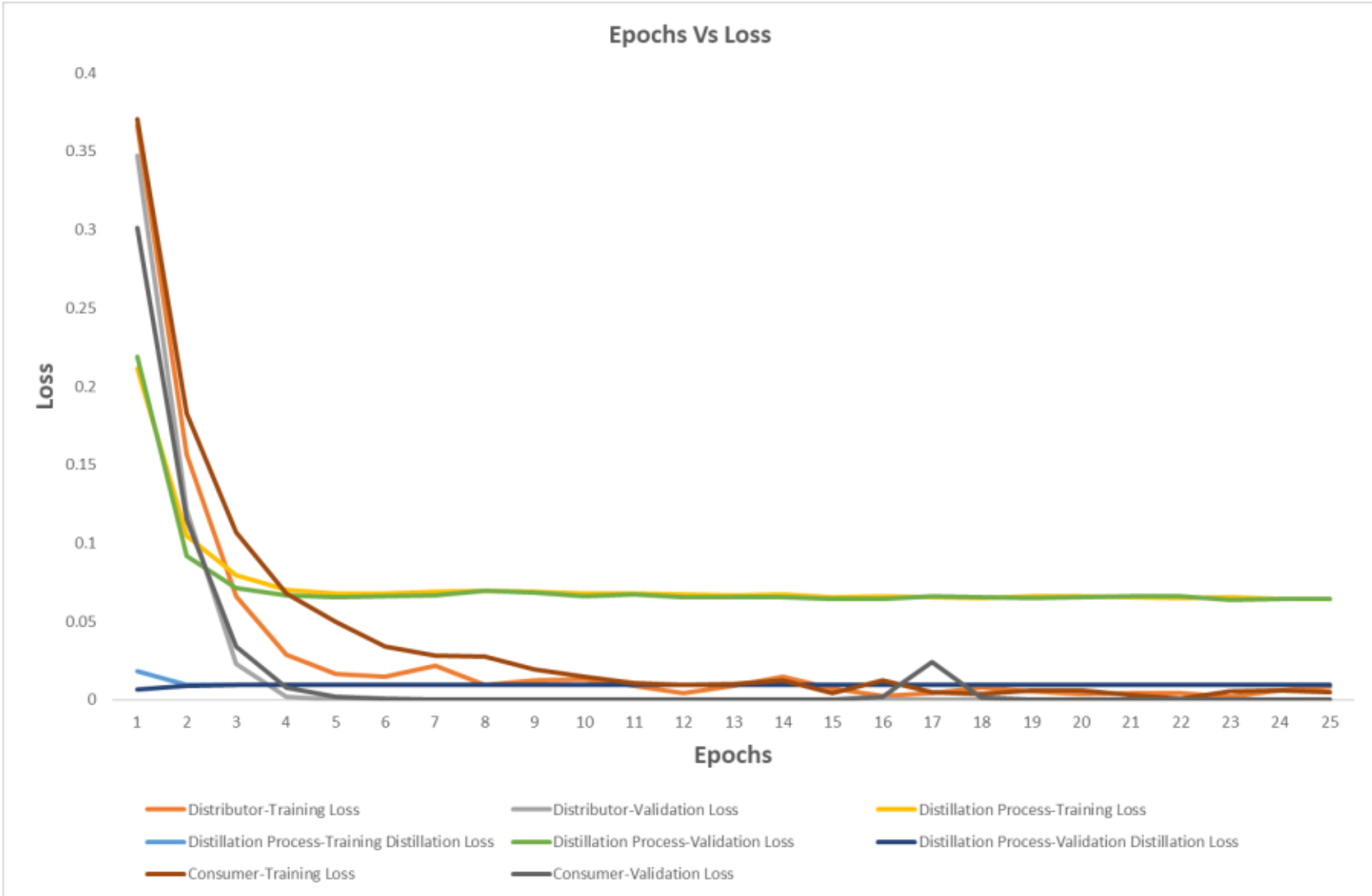

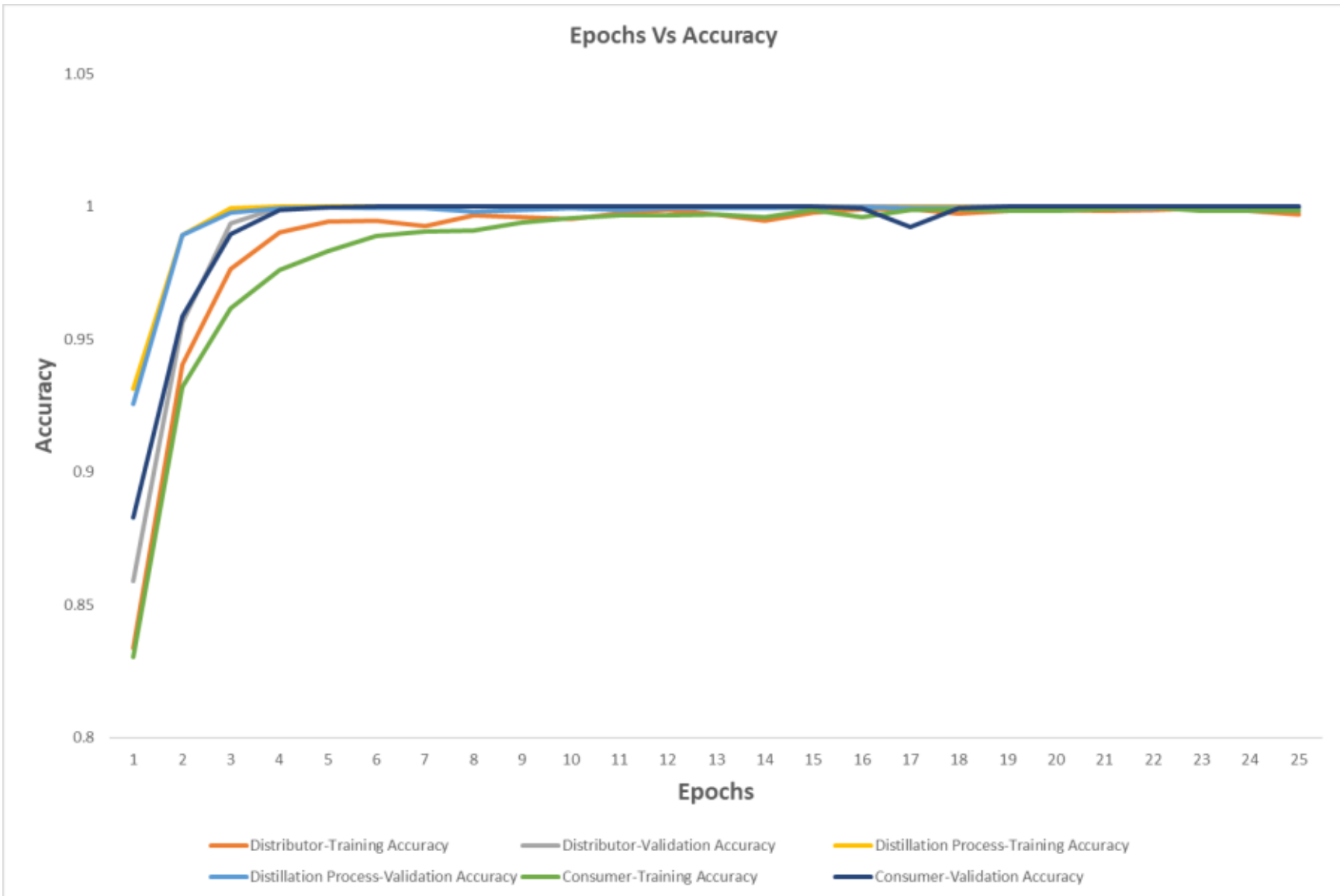

**Fig. 4.** Model Distillation MD training and performance results for networks

**Fine-tuning:** Fine tuning of the Whisper network is achieved using COMMON_VOICE_11_0 dataset for 4000 maximum steps, 500 warmup steps, learning rate at 1e-5, metric as WER, training batch size at 16, evaluation batch size at 8, datatype as floating point 16 and maximum length as 255.

Fine tuning of the M2M100 is achieved using IITB-ENGLISH-HINDI dataset for 20 epochs, decay of weight as 0.01, metric as BLEU, training batch size at 32, evaluation batch size at 64, learning rate at 2e-5, datatype as floating point 16 and maximum length as 128.

Fine tuning of the MMS is achieved using COMMON_VOICE_11_0 dataset for 10 epochs, 100 evaluation steps, learning rate at 1e-3, metric as WER, training batch size at 32, datatype as floating point 16 and maximum length as 255.

Table 2 and Figure 5, displays training for the Whisper, M2M100, and MMS-TTS and performance on respective dataset. Based on these results, it is worth noting that during PREPROCESSOR training of Whisper it takes around 10000 steps to stabilize the model. Whereas, in case of POSTPROCESSOR training of M2M100 and MMS-TTS, models stabilize around 10000 and 250 steps respectively.

**Table 2.** Training and performance results of PREPROCESSOR and POSTPROCESSOR.

| Model | Metric | Steps | S-Multiplier | T-Loss | V-Loss | Score |
|---|---|---|---|---|---|---|
| W/M/T | W/B/W | 100 | 100/100/1 | 0.2567/0.2765/4.905 | 0.3075/0.3273/2.3075 | 0.4463/0.5347/0.456 |
| W/M/T | W/B/W | 200 | 100/100/1 | 0.1967/0.2165/0.299 | 0.3558/0.3756/0.215 | 0.3313/0.5424/0.28 |
| W/M/T | W/B/W | 300 | 100/100/1 | 0.1125/0.1323/0.2659 | 0.3214/0.3412/0.167 | 0.3259/0.5871/0.232 |
| W/M/T | W/B/W | 400 | 100/100/1 | 0.0818/0.1016/0.2398 | 0.2519/0.2717/0.161 | 0.3201/0.6013/0.229 |
| W/M/T | W/B/W | 500 | 100/100/1 | 0.0312/0.051/0.127 | 0.1679/0.1877/0.156 | 0.321/0.6276/0.223 |
| W/M/T | W/B/W | 600 | 100/100/1 | 0.0108/0.0306/0.095 | 0.1455/0.1653/0.1455 | 0.285/0.6432/0.221 |
| W/M/T | W/B/W | 700 | 100/100/1 | 0.0051/0.0249/0.081 | 0.1251/0.1449/0.1251 | 0.2645/0.7338/0.224 |
| W/M/T | W/B/W | 800 | 100/100/1 | 0.0027/0.0225/0.0511 | 0.1995/0.2193/0.1995 | 0.2377/0.7674/0.217 |
| W/M/T | W/B/W | 900 | 100/100/1 | 0.0005/0.0203/0.027 | 0.1572/0.177/0.1572 | 0.2221/0.7824/0.203 |
| W/M/T | W/B/W | 1000 | 100/100/1 | 0.0002/0.0119/0.021 | 0.0573/0.0771/0.1573 | 0.2017/0.8915/0.197 |

## Legends - Table 2:

Model as W for Whisper, M for M2M100 and T for MMS-TTS. Metric as W for WER and B for BLEU. T-Loss as Loss of Training. S-Multiplier is multiplied to Steps to get total steps for respective model. V-Loss as Loss of Validation. Score is model performance on the specified metric for the model.

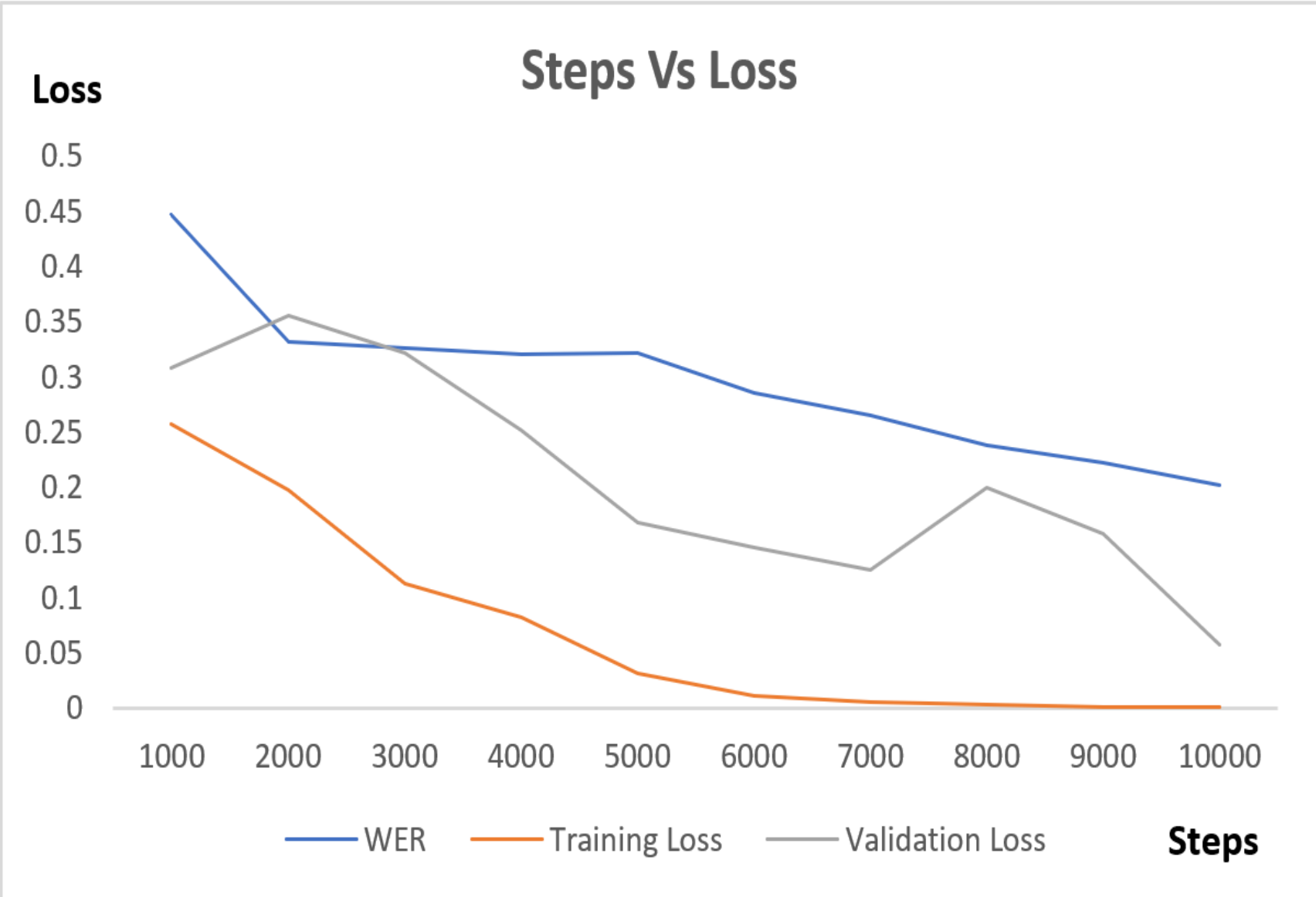
Steps Vs Loss
Loss
0.5
0.45
0.4
0.35
0.3
0.25
0.2
0.15
0.1
0.05
0
1000
2000
3000
4000
5000
6000
7000
8000
9000
10000
WER
Training Loss
Validation Loss
Steps

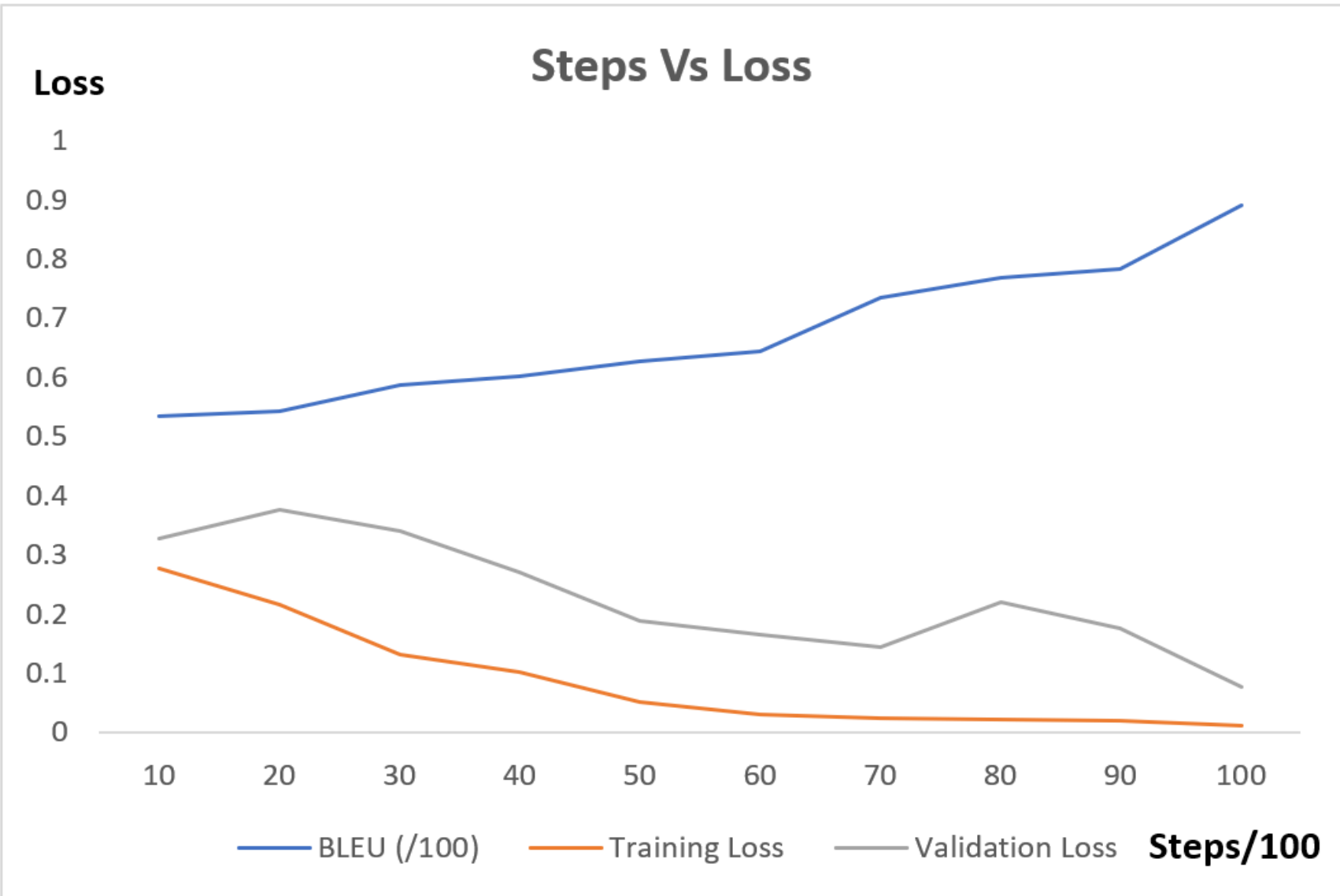
Steps Vs Loss
Loss
1
0.9
0.8
0.7
0.6
0.5
0.4
0.3
0.2
0.1
0
10
20
30
40
50
60
70
80
90
100
BLEU (/100)
Training Loss
Validation Loss
Steps/100

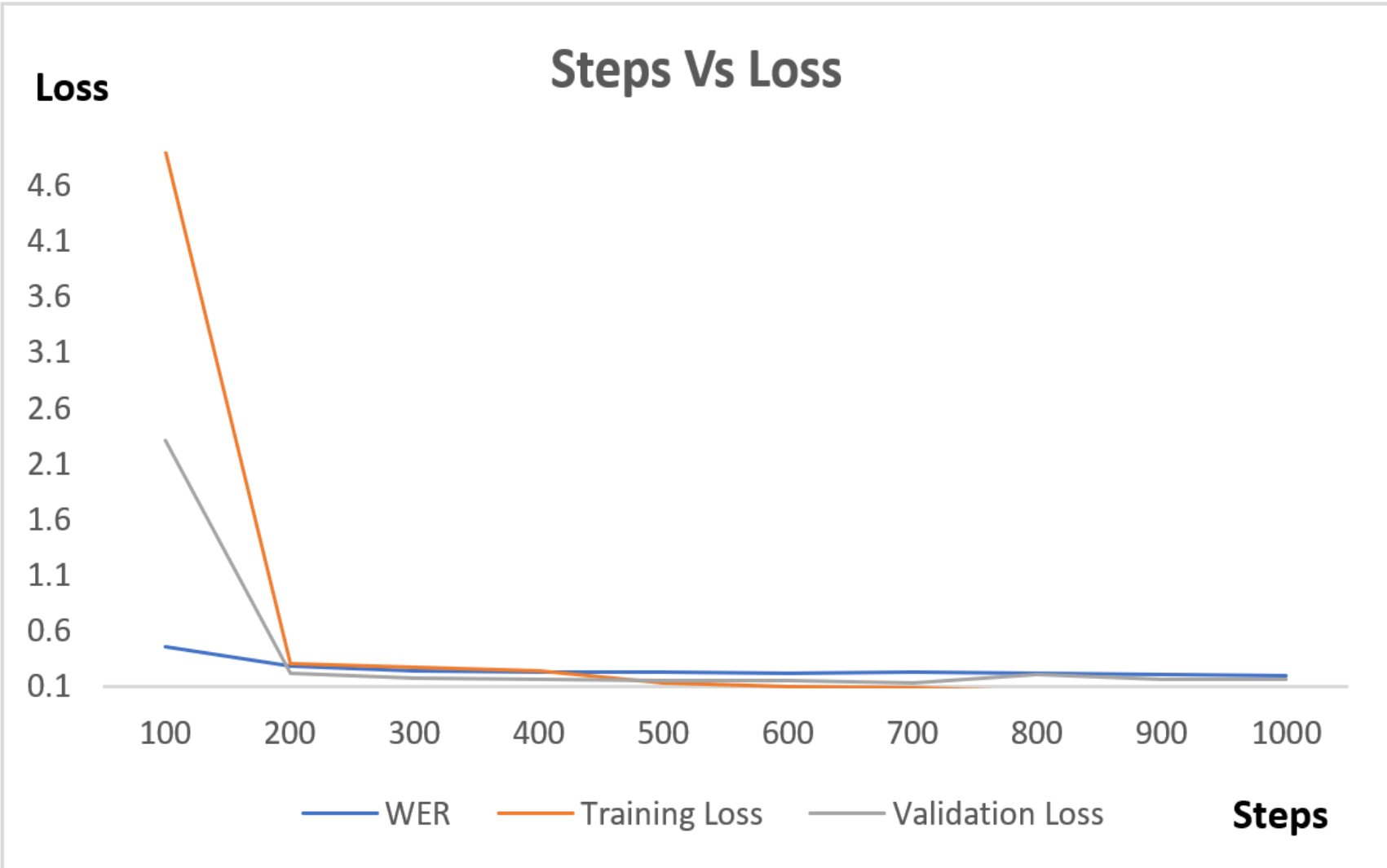

**Fig. 5.:** Training and performance results of PREPROCESSOR and POSTPROCESSOR

**Final Results:** Figure 6 displays the PDFTEMRA outputted inferences. As shown in Figure 6, when user interacts with application with input voice prompts of "Sar Dard" ("Headache pain") in Hindi (/English) and the network yields respective English Question Translation, English and Braille reply along with Hindi Audio responses.

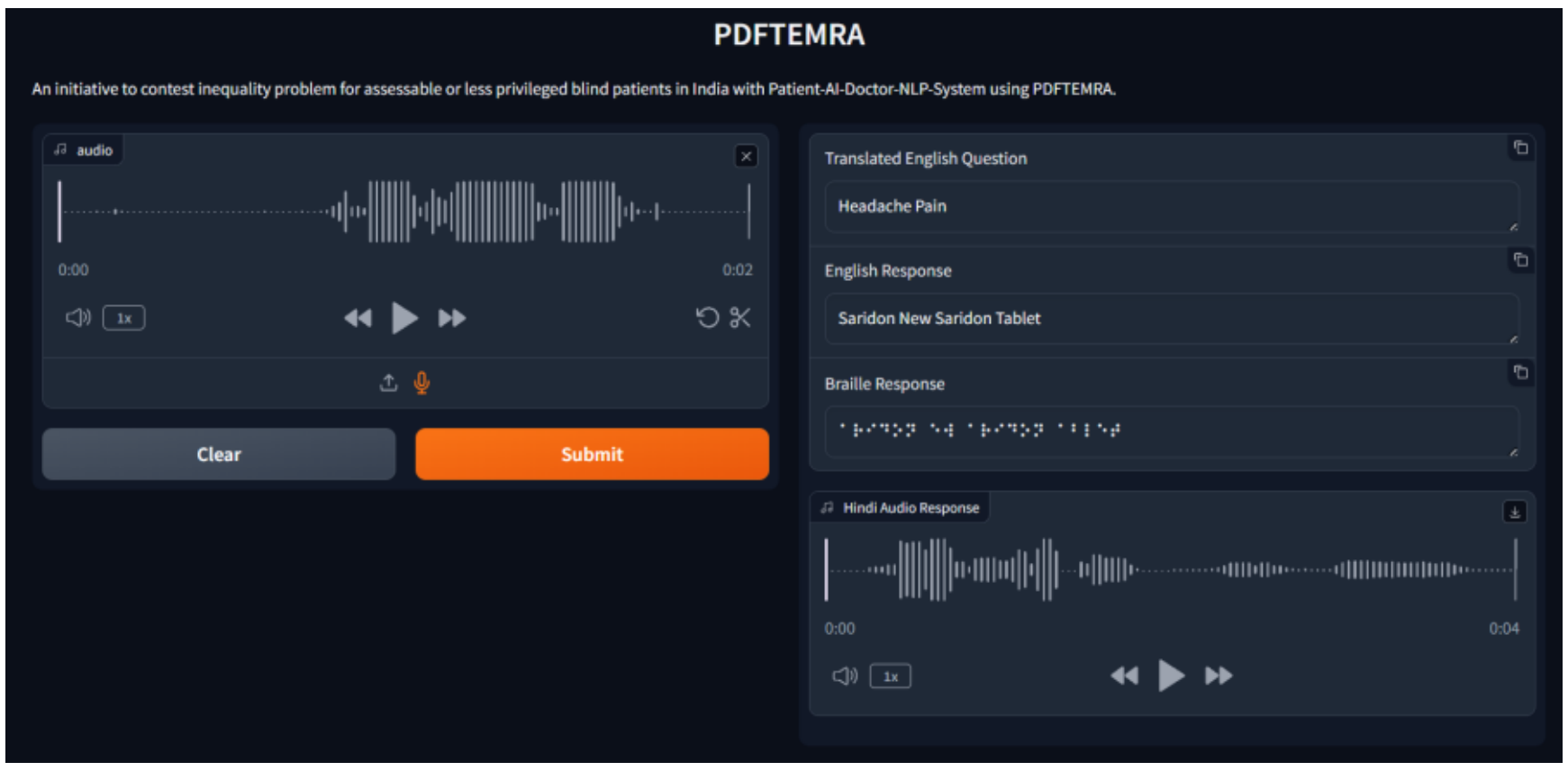

**Fig. 6.** Inference with PDFTEMRA.

# 5 Conclusions & Future Work

This work delivers an efficient and effective network, an artificial intelligence and natural language processing founded initiative, a Patient-Doctor-NLP-System for least privileged, PDFTEMRA, and a custom medical dataset, PADT. Multiple manifold advanced mathematical and engineering concepts helps to achieve comparable state-of-the-art network performance. Reader should note that, while executing, the results could be little different due to probabilistic nature. Achieving even better results the modifications like, different selection of the distributor network, hyper-parameter setups, better adaptation techniques, and overcoming experimentation hardware limitation of 20 CPU, 8GB VRAM GPU with 32 GB RAM. While this study nominated Hindi-English spoken-languages and optimized network and the hyper-parameter settings accordingly, the study and setup could be utilized for multiple native spoken languages across the world with minimal tweaks or optimizations to achieve similar or improved inferences.

# 6 Declarations

## 6.1 Competing interests

The corresponding author affirms that there is no conflict of interest on behalf of all authors. We (the authors) affirm that we do not have any opposing benefits to disclose that could be financial or personal relationship with any third party that can influence the article.